\def\year{2021}\relax
\documentclass[letterpaper]{article} 
\usepackage{aaai21}  
\usepackage{times}  
\usepackage{helvet} 
\usepackage{courier}  
\usepackage[hyphens]{url}  
\usepackage{graphicx} 
\usepackage{natbib}  
\usepackage{caption} 
\usepackage{booktabs}       

\usepackage{amsmath, amsfonts}
\usepackage{multirow}
\usepackage[table]{xcolor}
\usepackage{algpseudocode}
\usepackage{algorithm}

\newcommand*{\um}{UM}
\newcommand*{\ls}{\texttt{linear-skew} }
\newcommand*{\nn}{\texttt{nearest-neighbors} }

\title{Learning Prediction Intervals for Model Performance}
\author {

        Benjamin Elder,\textsuperscript{\rm 1}
        Matthew Arnold, \textsuperscript{\rm 1}
        Anupama Murthi \textsuperscript{\rm 1}
        Jiri Navratil \textsuperscript{\rm 1}\\
}
\affiliations {
    \textsuperscript{\rm 1} IBM Research \\
    benjamin.elder@ibm.com, marnold@us.ibm.com, anupama.murthi@ibm.com, jiri@us.ibm.com
}

\begin{document}
\maketitle

\begin{abstract}
Understanding model performance on unlabeled data is a fundamental challenge of developing, deploying, and maintaining AI systems. 
Model performance is typically evaluated using test sets or periodic manual quality assessments, both of which require laborious manual data labeling.
Automated {\it performance prediction} techniques aim to mitigate this burden, but potential inaccuracy and a lack of trust in their predictions has prevented their widespread adoption. 
We address this core problem of performance prediction uncertainty with a method to compute {\it prediction intervals} for model performance. 
Our methodology uses transfer learning to train an {\it uncertainty model} to estimate the uncertainty of model performance predictions. 
We evaluate our approach across a wide range of drift conditions and show substantial improvement over competitive baselines. 
We believe this result makes prediction intervals, and performance prediction in general, significantly more practical for real-world use.
\end{abstract}


\section{Introduction}

Knowing when a model's predictions can be trusted is one of the key challenges in AI today. 
From an operational perspective, understanding the quality of model predictions impacts nearly all stages in  the model lifecycle, including pre-deploy testing, deployment, and production monitoring.
From a social perspective, prediction trust impacts society's willingness to accept AI as it continues replacing human decision making in increasingly important roles. 

Techniques such as {\it performance prediction}~\cite{guerra2008, schat2020data, talagala2019} strive to automatically predict the performance of a model with no human intervention.
Unfortunately these techniques have not yet gained mainstream adoption, ironically enough, due to their potential unreliability and the resulting lack of trust in their predictions.
Performance predictors are often surprisingly accurate when a base model is predicting on data similar to what it has already seen in training and test. 
However, it is well known that model behavior can be extremely difficult to predict on previously unseen data~\cite{7298640, 8601309}. 
It is not reasonable to expect a performance prediction algorithm to perfectly predict a base model's behavior in these scenarios.  
It is, however, reasonable to ask a performance predictor to quantify the uncertainty of its predictions so the application (or end user) can take appropriate precautions.

This paper describes a technique for computing {\it prediction intervals} on {\it meta-model} based performance predictions, to convey the degree to which the performance prediction should be trusted.
Our technique uses {\it meta-meta-modeling} in a multi-task setting to train an {\it uncertainty model} to compute prediction intervals.
Our approach makes no assumptions about the base model or performance predictor, and can easily be applied in other settings.
The use of a separate {\it meta-meta-model} to perform the uncertainty quantification allows the simultaneous prediction of both aleatoric (data-driven) and epistemic (model-driven) uncertainty.

A key challenge for an uncertainty model is predicting the unseen - ie, data that is substantially different than anything the model has seen in train or test.
A simple cross validation or leave-one-out training is unlikely to produce sufficient feature-space drift to be informative in this regard.
Our technique trains for these extreme cases by (1) simulating various levels of drift ranging from mild to extreme, and (2) training on these drift scenarios using {\it external} data sets that are different from the base model training set but of the same modality.
This approach enables the uncertainty model to learn how the performance predictor behaves in a variety of challenging scenarios, and uses this information to help predict risk when challenging scenarios arise in production.
Due to its prevalence and commercial importance, we focus on tabular data, and therefore chose logistic regression and random forest base models. 

We evaluate our uncertainty model on four different performance predictors and compare the uncertainty model against four different model-free baseline algorithms.  In every scenario, our uncertainty model outperforms all baselines, often by a large margin.  Even without the use of an uncertainty model, our approach of using drift simulation for calibration yielded significant improvements over traditional baselines.


\section{Related work}
\label{sec:related}

Prior work exists on performance prediction~\cite{guerra2008, chen2019confidence, pmlr-v97-finn19a, redyuk2019learning, schat2020data, talagala2019}, however, to the best of our knowledge, none that assigns uncertainty bounds to their predictions.
The field of domain generalization also includes some related work, for which a good recent review can be found in \cite{hospedales2020metalearning}. 
For example, \cite{Li2018LearningTG} use a cross-domain meta-learning approach to model training similar our procedure, but applied to the base classification problem. 

Of relevance are numerous methods for estimating the uncertainty of predictions from machine learning models in general. 
Many of these, ranging from classical statistical methods to state-of-the-art deep learning (DL) models (\cite{gal2015theoretically, Kendall2017_whatuncertainty, KoenkerQuantileRegression78, nix1994estimating}, also see \cite{5966350} for a review), could be applied to meta-model based performance prediction.

There are well-established parametric methods for prediction intervals, see for example~\cite{geisser2017predictive}.
Methods implicitly learning the error distribution are also available, for example by incorporating a feature-dependent variance into the loss function for iterative training procedures~\cite{nix1994estimating}. 
Furthermore, there has been significant progress in constructing neural architectures which simultaneously output a classification and an uncertainty prediction~\cite{brosse2020lastlayer, kabir2019optimal, 5428779, NIPS2018_7936}.

Non-parametric methods such as the jackknife and bootstrap ~\cite{efron1979, doi:10.1080/00031305.1983.10483087}, and more recent variations~\cite{doi:10.1080/01621459.2017.1307116, vovk2018cross, papadopoulos2008, vovk2012, vovk2005} can estimate the uncertainty of a statistical prediction without assuming a particular error model, but rely on the assumption that the distribution of the unlabeled data is the same as the train data. 
Ensemble methods have been proposed for both traditional ML~\cite{dietterich2000ensemble, KWOK1990327} and DL models~\cite{492f6c68703a4b6d97bb8509d817d00f,  NIPS2017_7219, NIPS2018_8080}. 
The variance of the ensemble predictions can be used as a feature-dependent measure of uncertainty. We implement this strategy as one of our baselines (see Sec.~\ref{sec:methodology-baselines}). 
Bayesian approaches have been extended to non-parametric applications, including neural networks~\cite{BISHOP1997, 10.5555/3045118.3045290, neal2011mcmc} and a popular approximate version of Bayesian neural networks - the Monte-Carlo dropout approximation~\cite{gal2015, NIPS2017_6949} - also serves as a baseline in our work.

    
\section{Method}
\label{sec:method}

Our approach estimates prediction intervals for the performance of a black-box classification model on a pool of unlabeled data.
First, we use meta-modeling to predict the accuracy of the base classification model (\textit{performance prediction}). 
Second, a pre-trained \textit{uncertainty model} is used to estimate a prediction interval, which describes the probable range for the true value of the accuracy.


\subsection{Performance predictors}
\label{sec:method-performance_predictors}

As its name suggests, {\it performance prediction} is the problem of estimating the value of a performance metric that a machine-learning model will achieve for a given pool of unlabeled data (referred to here as the production set). 
This work focuses on classification accuracy, but the same methods could be applied to other metrics such as the F1 score or the error of a regression model. 
We treat the base model as a black box from which only the vector of predicted class probabilities for each sample is available. 
We developed two types of performance predictors (four variants in all) which we will use as the basis for our prediction intervals. 

The performance predictors that we use in this work each output a confidence score for each unlabeled data point in the production set. 
This score, between zero and one, is an estimate of the likelihood that the base model predicted the correct class label.
For the purposes of computing an aggregate accuracy score for the production set, we take the average of these confidences. 
This confidence averaging produced better estimates of the accuracy than making binary correct/incorrect predictions for each sample. 

The \texttt{confidence} predictor is a simple, binning-based procedure, which recalibrates the base model confidence score for the most likely class~\cite{ZADROZNY2001}. 
The values of this confidence on the test set are gathered into a histogram (binned in increments of 0.1), and the base model accuracy is computed for each bin. 
A performance prediction for a data point is given by the average accuracy of the bin that spans that point's base model confidence.

Our \texttt{meta-model} performance predictor uses its own model to predict data points that are likely to be mislabeled by the base model. %
Training data for the \texttt{meta-model} predictor is created by relabeling the test set with binary labels indicating whether the base model correctly classified each sample. 
The meta-model, which is an ensemble of a Gradient Boosting Machine (GBM) and a logistic regression model, classifies each sample as correct or incorrect, and the probability assigned to the "correct" class is returned as the performance predictor confidence score. 
Further details of the \texttt{meta-model} predictor implementation are provided in the supplementary material.


\subsection{Uncertainty model}
\label{sec:method-uncertainty_model}

We propose a new approach for computing prediction intervals by using an \textit{uncertainty model} (a meta-meta-model) that learns to predict the behavior of a performance predictor (a meta-model). 
This uncertainty model (\um{}) is a regression model that quantifies the uncertainty of the performance predictor’s estimate of the base model accuracy on the production set. 
We pre-train the \um{} using a library of training datasets. 
The \um{} can then observe the behavior of a performance predictor on a new target dataset and generate a prediction interval.

The \um{}  must be trained using examples of performance prediction errors.
Each training sample for the \um{} consists of a full drift scenario, comprising: (1) labeled train and test datasets, (2) a pool of unlabeled data (the production set), (3) a base classification model trained on the train set, and (4) a performance predictor trained for this dataset and base model.
We generate a large number of such examples using the simulation procedures described in Sec.~\ref{sec:method-drift}.
The target values are the (absolute) differences between the true and predicted accuracy on the entire production set. This method could be extended to predict the signed value of the errors, allowing for asymmetric prediction intervals.

The \um{} architecture is an ensemble of GBM models. 
Experimentally, we found that an ensemble of ten models reduced prediction variance and led to improved accuracy. 
The model was trained using a quantile loss function, which naturally enables the calculation of prediction intervals targeted to capture the true error with a specified probability.
Further implementation details are described in the supplementary material.


\paragraph{Features}

\begin{table*}[t]
\footnotesize
\centering
  \begin{tabular}{ | l | l | l | c | c |}
    \hline
    Source & Name & Description & Type & Count \\ \hline
    Base & top confidence & confidence score of predicted class & D & 3 \\ \cline{2-5}
    & (top-2nd) confidence & top - second highest class probability & D & 3\\ \cline{2-5}
    & confidence entropy & entropy of confidence vector & D & 3\\ \cline{2-5}
    & class frequency & relative frequency of predicted classes  & D & 3\\ \cline{2-5}
    & entropy ratio & avg. prod. set entropy/avg. test set entropy & P & 1\\ \cline{2-5}
    & bootstrap & size of bootstrap conf. intervals for avg. accuracy & N & 1\\ \hline
    \hline
    Perf. Pred. & predicted change & predicted prod. acc. - base test acc. & P & 1\\ \cline{2-5}
    & avg. pred. stdev. & stdev. of confidence scores (prod) & P & 1\\ \cline{2-5}
    & avg. pred. entropy & entropy of confidence histogram (prod) & P & 1\\ \cline{2-5}
    & predicted uncertainty & intrinsic uncertainty interval & I & 1\\ \cline{2-5}
    & bootstrap & size of bootstrap conf. intervals for pred. acc. & N & 1\\ \cline{2-5}
    & ensemble whitebox & whitebox stats from meta-model ensemble & I & 2\\ \cline{2-5}
    & calibration whitebox & whitebox stats from calibration & I & 2\\ \cline{2-5}
    & gbm whitebox & internal stats from gbm meta-model & I & 16\\ \hline
    \hline
    Proxy & top confidence & confidence score of predicted class & D & 9\\ \cline{2-5}
    & (top-2nd) confidence & top - second highest class probability & D & 9\\ \cline{2-5}
    & best feature & projection onto most important feature & D & 3\\ \cline{2-5}
    & num import. feat. & num. features to make 90\% feat. importance & I & 1\\ \hline
    \hline
    Drift & accuracy & test vs. prod classification accuracy & P & 3\\ \cline{2-5}
    & (top-2nd) confidence & top - second highest class probability & D & 9\\ \hline
    \hline
    Other & PCA projection & projection onto highest PCA component & D & 3\\ \hline
  \end{tabular}
  \caption{\label{table:feature-description}\footnotesize Source models, names, descriptions, types (D=\texttt{Distance}, P=\texttt{Prediction}, N=\texttt{Noise}, I=\texttt{Internal}), and counts for all \um{} features.}
\end{table*}

The \um{} was trained using a set of derived features that are generic enough to be compatible across datasets with varying feature spaces and numbers of classes.
The derived features are extracted from a number of models, including (1) the base classification model, (2) the performance predictor, (3) a group of proxy models, and (4) a group of drift models. 
The proxy models (one logistic regression, one random forest, and one GBM) were trained on the same features and classification task as the base model, and provide a complementary perspective on the classification difficulty of each data point. 
The drift models are random forest models trained to predict whether a given sample came from the test set or the production set. 
They provide direct insight into the degree of feature space drift for a given scenario. 
Further implementation details for the proxy and drift models are provided in the supplementary material.

A full list of the features used for the \um{} is shown in Table~\ref{table:feature-description}.  
The procedure for constructing the features of type \texttt{Distance} starts by choosing a function $f$ that maps any feature vector to a scalar value, for example the highest value from the base model confidence vector.
The value of this function is used to construct two histograms, one for the samples in the test set, and one for the production set. 
Finally, the distance between these two histograms is computed using some distance function $D$. 
We used three functions for $D$: the Kolmogorov–Smirnov metric $ D_1 = \max_i\vert P_i - Q_i\vert$, the inverse overlap $D_2 = \sum_i \max (p_i - q_i, 0)$, and the squared inverse overlap $D_3 = \sum_i \big(\max (p_i - q_i, 0)\big)^2$. 
Here $p$ and $q$ are the normalized histograms from the test and production sets, $i$ indexes the bins in the histograms (which must be identically spaced), and $P,Q$ are the CDFs corresponding to $p,q$.

The remaining features can be grouped into three types: \texttt{Prediction}, \texttt{Noise}, and \texttt{Internal}.
The \texttt{Prediction} features are directly derived from the predictions of one of the source models (without using the \texttt{Distance} procedure). 
These include the entropy of the base model predictions, the change in accuracy predicted by the performance predictor, and the accuracy of the drift classifiers. 
The \texttt{Noise} features are the size of bootstrap confidence intervals for the average of the top base model confidence and the performance predictor confidence, both of which approach zero as the number of points in the production set approaches infinity.
The \texttt{Internal} features are white-box quantities extracted from the performance predictor or proxy models, such as the {\it intrinsic} prediction intervals described in Sec.~\ref{sec:methodology-baselines}, or the difference between the calibrated and uncalibrated performance predictions.
A complete description of the features listed here is given in the supplementary material. 

An ablation study is presented in Table~\ref{table:features-importance}, showing the performance of the \um{}, coupled with the \texttt{meta-model} performance predictor, obtained using subsets of the full set of features described above. 
This study shows that the most effective information for understanding the performance prediction uncertainty comes from the drift models. 
It also shows that including the more numerous and computationally expensive \texttt{Distance} and \texttt{Internal} features deliver a significant performance boost over the simpler \texttt{Prediction} and \texttt{Noise} features. 

\begin{table}
\footnotesize
  \begin{tabular}{ | l | c | c | c | c | c |}
  \hline
    &\multicolumn{5}{c |}{Average Cost ($\alpha=0.9$)} \\  \cline{2-6}
    & Base & Perf. Pred. & Proxy & Drift & All \\ \hline
    \texttt{Distance} & 1.46 &\cellcolor{gray!25} -- & 1.21 & 1.18 & 1.07 \\ \hline
    \texttt{Internal} & 1.61 & 1.15 &\cellcolor{gray!25} -- &\cellcolor{gray!25} -- & 1.05 \\ \hline
    \texttt{Prediction} &\cellcolor{gray!25} -- & 1.29 & 1.87 &\cellcolor{gray!25} -- & 1.33 \\ \hline
    \texttt{Noise} & 1.46 & 1.59 &\cellcolor{gray!25} -- &\cellcolor{gray!25} -- & 1.32 \\ \hline
    All & 1.30 & 1.25 & 1.21 & 1.16 & 1.0 \\ \hline
  \end{tabular}
  \caption{\label{table:features-importance} Average uncertainty model cost, Eq.~\eqref{eq:cost_function}, for the \texttt{meta-model} predictor, with $\alpha=0.9$, normalized by the value when all features are used. }
\end{table}

   
\section{Experimental methodology}
\label{sec:methodology}

\begin{figure*}[h]
	\centering
	\includegraphics[width=0.9\textwidth]{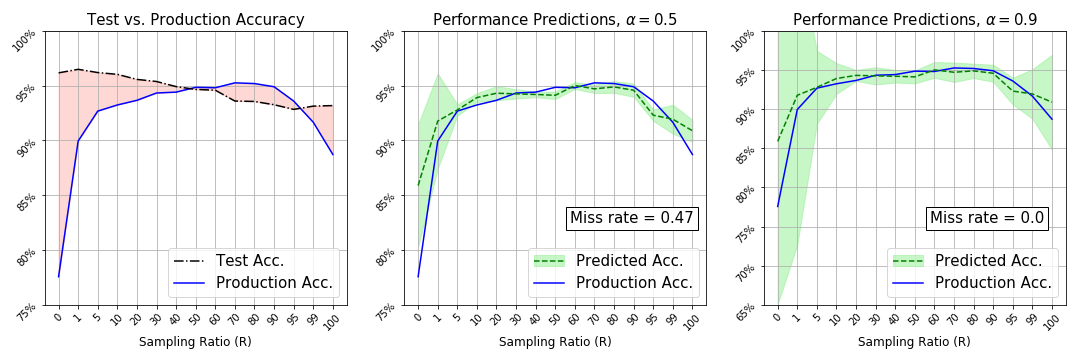}
	\caption{\label{fig:performance-prediction-example}\footnotesize Example of performance prediction and the \um{} for \ls drift scenarios. The left plot shows the model accuracy drift (shaded area) induced by Alg.~\ref{alg:biased-splits}. The middle (right) plot shows the accuracy predicted by the \texttt{meta-model} predictor, with \um{} prediction intervals calibrated to $\alpha = 0.5$ ($\alpha=0.9$).}
\end{figure*}

Fig.~\ref{fig:performance-prediction-example} shows example results for \ls drift scenarios (Sec. \ref{sec:method-drift}) simulated from the bng-zoo dataset. 
The middle and right panel show accuracy predictions from the \textit{meta-model} performance predictor, and the uncertainty model prediction intervals calibrated using two different levels of $\alpha$, (0.5 and 0.9), as described in Sec. \ref{sec:methodology-calibration}.


\subsection{Drift simulation}
\label{sec:method-drift}

Training and evaluating the \um{} requires examples of data drift. The breadth of the drift examples used for training largely determines the quality and coverage of the resulting model.
Tabular labeled datasets containing sufficient naturally occurring drift are difficult to obtain, therefore we chose to generate such examples through resampling-based simulation. 
%
%
We focus on \textit{covariate shift} because it has been shown to encompass a wide range of real-world drift scenarios~\cite{card-smith-2018-importance}. 

We use two different algorithms for generating drift: (1) \ls, is designed for breadth of coverage, for providing training data, and ensuring the generation of extreme drift (2) \texttt{nearest-neighbors} is designed to simulate drift more likely to occur in the real world for an additional evaluation scenario.
Examples of drift generated by both algorithms is included in the supplementary material.

\paragraph{Linear-skew} 
The \ls method requires choosing a feature dimension $(F)$ along which the bias will be induced, and a threshold $t$ to split the dataset into two buckets. 
The sampling parameter $R$ controls the ratio of sampling from the two buckets for the train/test sets and the production set, thus also controlling the amount of drift in the scenario. 
When $R = 50$, the train, test, and production sets all have the same distribution, and there is no drift. 
When $R=0$ or $R=100$, there is no overlap between the train/test distribution and the production distribution. 
The \ls procedure is described in Alg.~\ref{alg:biased-splits}.

\begin{algorithm}[h]
  \caption{Algorithm to create \ls drift scenarios}
  \label{alg:biased-splits}
  \footnotesize
\begin{algorithmic}
    \State {\bfseries Input:}
    Dataset $X \subset \mathcal{X}$; $p_{tr},p_{te},p_{pr}\in [0,1]: p_{tr}+p_{te}+p_{pr}=1$; 
    Feature dimension $F: \mathcal{X}^{(F)}\subset \mathcal{X}$; threshold function $t: \mathcal{X}^{(F)} \rightarrow \{X_A,X_B\}$; Sampling ratio $R \in [0,100]$; minibatch size $b$
    \State {\bfseries Output:} $X_{tr}, X_{te}, X_{pr} \subset X$
    \State $X_A, X_B, X_{tt}, X_{pr} \leftarrow \{\}$
    \For{$x$ {\bfseries in} $X$}
    \State Add $x$ to $t(X^{(F)}) \in \{X_A, X_B\}$ \Comment{Add data point to bucket, defined by threshold $t$}
    \EndFor
    \While{$|X_A| > b$ {\bfseries{and}} $|X_B| > b$} \Comment{Randomly sample $X_{tt}$, $X_{pr}$ from buckets until out of data points}
        \State Add $(p_{tr}+p_{te})\times \frac{R}{100}\times b$ points from $X_A$ and $(p_{tr}+p_{te})\times (1-\frac{R}{100}) \times b$ points from $X_B$ into $X_{tt}$
        \State Add $p_{pr}\times (1-\frac{R}{100}) \times b$ points from $X_A$ and $p_{pr}\times \frac{R}{100} \times b$ points from $X_B$ into $X_{pr}$
    \EndWhile
    \State Randomly split $X_{tr}, X_{te} \leftarrow X_{tt}$ with proportions $\frac{p_{tr}}{p_{tr}+p_{te}}$ and $\frac{p_{te}}{p_{tr}+p_{te}}$
\end{algorithmic}
\end{algorithm}

\paragraph{Nearest-neighbors} 

The \nn algorithm strives to simulate a particular demographic either appearing, or disappearing from production traffic. 
It does so by sampling a data point and then uses nearest neighbors to identify other data points that are similar (nearest neighbors) or dissimilar (furthest neighbors) and remove them from the dataset. 
We used this algorithm to create fairly severe drift, removing 50-70\% of the original data points, which we believe covers the range of realistic possible drift.
The \nn algorithm is described in Alg.~\ref{alg:nearest-neighbors}.

\begin{algorithm}[h]
  \caption{Algorithm to create \nn drift scenarios}
  \label{alg:nearest-neighbors}
\begin{algorithmic}
    \footnotesize
    \State {\bfseries Input:}
    Dataset $X \subset \mathcal{X}$; $p_{tr},p_{te},p_{pr}\in [0,1]: p_{tr}+p_{te}+p_{pr}=1$; $P_{set}\in [0,1]$; $P_{near}\in [0,1]$; $P_{down}\in [0,1]$
    \State {\bfseries Output:} $X_{tr}, X_{te}, X_{pr} \subset X$
    \State Randomly split data into $X_{pr}$ and $X_{tt}$ with proportions $p_{pr}$ and $1-p_{pr}$
    \State With probability $P_{set}$, set downsample set $X_{down}=X_{tt}$ and $X_{rand}=X_{pr}$, else $X_{down} = X_{pr}$ and $X_{rand}=X_{tt}$ \Comment{Choose distribution to bias non-randomly}
    \State Choose point $p \in X_{down}$ at random
    \State Order points $x\ne p \in X_{down}$ by distance from $p$
    \State Choose $D$=nearest $(N)$ with probability $P_{near}$ else $D$=furthest $(F)$ \Comment{Choose nearest or furthest bias}
    \State Remove the fraction $P_{down}$ points which are $D \in \{N,F\}$ from $p$
    \State Remove fraction $P_{down}$ from $X_{rand}$ randomly \Comment{Randomly downsample non-biased distribution}
    \State Randomly split $X_{tr}, X_{te} \leftarrow X_{tt}$ with proportions $\frac{p_{tr}}{p_{tr}+p_{te}}$ and $\frac{p_{te}}{p_{tr}+p_{te}}$
\end{algorithmic}
\end{algorithm}


\subsection{Datasets and settings}

For our experiments we use a set of fifteen publicly available tabular datasets, sourced from Kaggle, OpenML, and Lending Club: 
Artificial Character, Bach Choral, Bank Marketing, BNG Zoo, BNG Ionosphere, Churn Modeling, Creditcard Default, Forest Cover Type, Higgs Boson, Lending Club (2016 Q1, 2017 Q1), Network Attack, Phishing, Pulsar, SDSS, and Waveform. 
Details of the individual dataset's characteristics and our pre-processing procedures are provided in the supplementary material. 

To simulate the effect of training the \um{} on an offline library of training datasets and then deploying it to make predictions on a new, unseen target dataset, we conducted our experiments in a leave-one-out manner. 
Each dataset was chosen in turn as the target, and its \um{} was trained on the remaining training datasets. 
All results are averaged over these fifteen different \um{}s.

Since we focused on tabular data, we used random forest and logistic regression base models for all experiments. 
In the \ls simulations we chose two features per dataset, and performed Alg.~\ref{alg:biased-splits} for each feature with fifteen values of the sampling ratio $R = 0, 1, 5, 10, 20, 30, 40, 50, 60, 70, 80, 90, 95, 99, 100$. 
This was repeated using five random seeds, giving a total of 300 drift scenarios per dataset.\footnote{Except for the Network Attack dataset, which only has one feature amenable to this procedure, see supplementary material.}
For the \nn simulations, 300 scenarios were generated for each dataset/base model combination with parameters $P_{set}=0.5$, $P_{near}=0.5$, and $P_{down}\in [0.5,0.7]$, for a total of 9000 scenarios. 


\subsection{Model-Free Baselines}
\label{sec:methodology-baselines}

A set of model-free baseline techniques for computing prediction intervals are compared against the UM. 
The first set of techniques are three {\it intrinsic} methods which leverage white-box information from the performance predictors.
These intrinsic methods produce uncertainty estimates for each point in the production set, and the average of these estimates is used as the (uncalibrated) prediction interval.

For the \texttt{confidence} predictor, if the accuracy in bin $k$ is $a_k$ and the number of samples falling into bin $k$ is $n_k$, we compute an uncertainty score for each point in the $k-$th bin as 
$u_k = \sqrt{a_k(1-a_k) /n_k}$, 
which is the standard error of a Bernoulli distribution with parameter $a_k$. 

For the \texttt{meta-model} predictor, we created two variants that replace the GBM and logistic regression meta-models with different classifiers.
The \texttt{crossval} predictor uses an ensemble of ten random forest models, each trained with a different cross-validation fold of the test set, and the standard deviation of their predictions is used as the uncertainty estimate. 
The \texttt{dropout} predictor uses an XGBoost~\cite{10.1145/2939672.2939785} model with the DART~\cite{pmlr-v38-korlakaivinayak15} booster, which applies dropout~\cite{dropout} regularization to GBM models. 
In the spirit of the Monte-Carlo dropout approach for Bayesian neural networks~\cite{gal2015, NIPS2017_6949}, ten predictions are made for each sample with dropout turned on to introduce randomness in the confidence scores, giving an estimated average and standard deviation for the model accuracy. 

Besides the intrinsic baselines, we also compare with three other baseline methods, which produce prediction intervals using: (1) the standard error of the mean of the performance predictor confidences, (2) the size of a bootstrap uncertainty interval for the mean of these confidences, and (3) a constant sized prediction interval.


\subsection{Evaluation Metric}
\label{sec:methodology-metric}

Evaluating the quality of a set of prediction intervals involves a trade-off between two opposing kinds of errors:  prediction intervals that are too small and do not capture the magnitude of the true error (Type I cost), and prediction intervals that are unnecessarily large (Type II cost). 
In an ad hoc comparison between two methods generating prediction intervals, it is common that one method does not dominate the other in the sense of having both smaller Type I and Type II error. 
A comprehensive comparison of such methods requires making a tradeoff between the two.

One common approach to quantifying this trade-off is to scale each set of prediction intervals by a constant factor to achieve a common miss rate (eg 5\%), and then compare their average size or average excess. 
We chose instead to evaluate results based on a cost function which penalizes both types of error: 
\begin{align}
\begin{split}
\label{eq:cost_function}
C_\alpha(\vec{\delta}, \vec{u}) = \sum_i &\bigg[ \alpha \max \big( \delta_i - u_i, 0\big)\\ 
&\ \ \ \  + (1-\alpha) \max \big( u_i - \delta_i,  0\big) \bigg] \, . 
\end{split}
\end{align}
In Eq.~\eqref{eq:cost_function}, $\delta_i = \vert a_i - p_i\vert$ is the difference between the true base model accuracy and the performance prediction for production set $i$, and $u_i$ is the (single-directional) size of the scaled prediction interval. 
The cost function parameter $\alpha \in [0,1]$ can be adjusted to control the balance between the two types of error.


\subsection{Calibration}
\label{sec:methodology-calibration}

Prediction intervals must be calibrated in order to achieve reliable performance for a scale-sensitive metric such as Eq.~\eqref{eq:cost_function}. 
We compare two calibration methods, one with external drift scenarios as a holdout set, and one without. 
Without the holdout set for calibration (Target Calibration), we used the approximation that the performance prediction error is normally distributed, and multiplied the uncalibrated prediction intervals by the Z-score of the desired confidence interval (determined by the cost function parameter $\alpha$).\footnote{The cost is minimized by balancing the two terms in Eq.~\eqref{eq:cost_function}, thus the $\alpha$\%-confidence interval chosen for calibration. For example, if $\alpha=0.95$ is chosen, then the scale factor is $Z=1.96$. }
This calibration method is applied to all of the model-free baselines described in Sec.~\ref{sec:methodology-baselines}. 

The second calibration method (TL Calibration) used the simulated drift scenarios from external holdout datasets as a calibration set. 
A constant scale factor was computed which minimized the cost function Eq.~\eqref{eq:cost_function} for each set of prediction intervals on this hold-out set. 
This calibration method was applied to the model-free baselines using 100\% of the holdout drift scenarios, and to the \um{} prediction intervals using an 80\%/20\% train/test split of the drift scenarios.


\section{Experimental results}
\label{sec:results}

In this section we demonstrate the performance of our \um{}, trained using the \ls drift scenarios, and evaluated using both the \ls and the \nn style scenarios. 
We use the four performance predictors described in Sec.~\ref{sec:method-performance_predictors}, and the four baselines described in Sec.~\ref{sec:methodology-baselines}.  
The quality of the prediction intervals is measured by the cost function $C_\alpha$ from Eq.~\eqref{eq:cost_function}, with $\alpha$ between 0.5 and 0.95. 
This range covers most reasonable user preferences for penalizing under- vs. over-shooting of the appropriate prediction interval size.

\paragraph{Linear-skew results}
\label{sec:results-same-scenario-training}

\begin{figure}[t!]
    \centering
    \includegraphics[width=0.8\columnwidth]{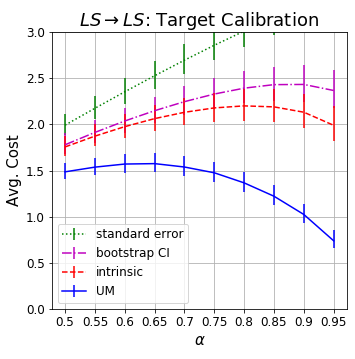}
    \includegraphics[width=0.8\columnwidth]{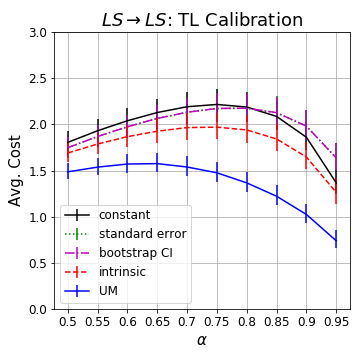}
\caption{\label{fig:biased-target-results} Average cost (Eq.~\eqref{eq:cost_function}) for prediction intervals. The baselines from Sec.~\ref{sec:methodology-baselines}, calibrated without (top) and with (bottom) external datasets, are compared to the \um{}. The evaluation uses the \ls drift scenarios. }
\end{figure}

%

Fig.~\ref{fig:biased-target-results} compares the cost $C_\alpha$ of the \um{} and the standard error, bootstrap, and intrinsic baselines described in Sec.~\ref{sec:methodology-baselines}, evaluated using the \ls drift scenarios. 
The results are averaged across the four performance predictors\footnote{There are only three sets of results for the ``intrinsic" curve, as the \texttt{meta-model} predictor has no intrinsic prediction interval.}, and calibrated using the target dataset method (top) and external holdout drift scenarios (bottom), as described in Sec.~\ref{sec:methodology-calibration}. 
The error bars in Fig.~\ref{fig:biased-target-results} indicate the 95\% bootstrap confidence interval. 
The bottom plot also includes the calibrated constant baseline. 

It is clear from the upper panel of Fig.~\ref{fig:biased-target-results} that the \um{} method trained with the \ls simulated drift scenarios substantially outperforms the baselines. 
This is especially true for moderate to high values of $\alpha$, which correspond to penalizing prediction intervals that are too small more than intervals which are too large. 
Comparing this result to the bottom panel, we see that merely calibrating with the simulated drift scenarios can dramatically improve the performance of the baseline methods. 
However, the \um{} still provides a major improvement at all values of $\alpha$.

\paragraph{Nearest-neighbors results}

\begin{figure}[t!]
    \centering
    \includegraphics[width=0.8\columnwidth]{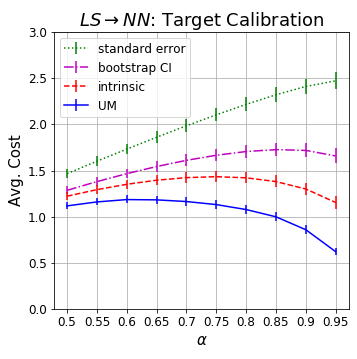}
    \includegraphics[width=0.8\columnwidth]{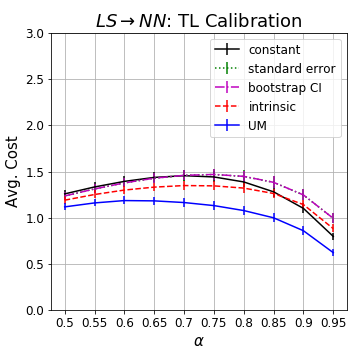}
    \caption{\label{fig:nn-TL-results} Same experiments as Fig.~\ref{fig:biased-target-results}, except using \nn drift scenarios for evaluation. }
\end{figure}

%

Fig.~\ref{fig:nn-TL-results} shows the same experiments as Fig.~\ref{fig:biased-target-results}, but using \nn scenarios for evaluation (and still using \ls scenarios for training).
The \um{} still outperforms the baselines for the full range of $\alpha$. 
The overall costs tend to be smaller, since the model accuracy drift in the \nn scenarios was smaller on average than in the \ls scenarios. 
This result confirms that the deliberately engineered \ls scenarios are able to provide effective training for more realistic, organically created \nn drift scenarios.

Table~\ref{table:results} provides all of the results in numeric form, broken down by performance predictor, and averaged across the same range of $\alpha$ values.
This confirms that the individual predictor results align with the averaged results previously shown.

\paragraph{Generalization and Limitations}

Choosing the source datasets for this transfer-learning based approach is an important consideration. 
For a domain specific application, it is obviously preferable to choose source datasets from the same or a closely related domain. 
In addition to the domain, we expect that it is valuable to approximately match other dataset characteristics such as number of classes, number of features, feature sparsity, etc. 

For applications to other data modalities, for example image or text data, many standard benchmark datasets such as ImageNet could be used for pre-training. 
The models used in the \um{} and the feature extraction would need to be replaced or supplemented with modality appropriate architectures to extract meaningful features from the data. 
Finally, the base models used in the drift scenarios for training the \um{} should include models of the same class as those to which it will be applied at prediction time.

\section{Conclusion}
\label{sec:conclusion}

Performance prediction is an invaluable part of the deploying, monitoring, and improving an AI model. 
This paper addresses this problem by describing a novel technique for quantifying model uncertainty. It leverages multi-task learning and meta-meta-modeling to generate prediction intervals on any model-based performance prediction system. 
Our method substantially outperforms a group of competitive baselines on dataset shift produced by two different simulation mechanisms. 
We believe this work helps make performance prediction more practical for real-world use, and may encourage further innovation in this important area.

\begin{table}
\footnotesize
\centering
\begin{tabular}{| l | c | c | c | c |}
	\hline
	\multicolumn{5}{| c |}{LS (train) $\rightarrow$ LS (eval)} \\ \hline
	& \multicolumn{4}{| c |}{Predictor} \\ \hline
	Method & Confidence & Crossval & Dropout & Meta-model \\ \hline
	SE & 2.88 & 2.57 & 3.09 & 2.41 \\ \hline
	BS & 2.38 & 2.06 & 2.50 & 1.89 \\ \hline
	I & 2.23 & 1.72 & 2.20 & \cellcolor{gray!25} --  \\ \hline
	SE (TL) & 2.13 & 1.91 & 2.13 & 1.77 \\ \hline
	BS (TL) & 1.13 & 1.91 & 2.13 & 1.77 \\ \hline
 	C (TL) & 2.11 & 1.88 & 2.10 & 1.83 \\ \hline
	I (TL) & 1.83 & 1.67 & 1.86 & \cellcolor{gray!25} --  \\ \hline
	UM & $\mathbf{1.46}$ & $\mathbf{1.24}$ & $\mathbf{1.38}$ & $\mathbf{1.33}$ \\ \hline
    
	\hline
	\multicolumn{5}{| c |}{LS (train) $\rightarrow$ NN (eval)} \\ \hline
	& \multicolumn{4}{| c |}{Predictor} \\ \hline
	Method & Confidence & Crossval & Dropout & Meta-model \\ \hline
	SE & 1.97 & 1.87 & 2.48 & 1.75 \\ \hline
	BS & 1.59 & 1.41 & 2.00 & 1.29 \\ \hline
	I & 1.39 & 1.14 & 1.48 & \cellcolor{gray!25} -- \\ \hline
	SE (TL) & 1.36 & 1.25 & 1.58 & 1.15 \\ \hline
	BS (TL) & 1.36 & 1.25 & 1.58 & 1.15 \\ \hline
 	C (TL) & 1.30 & 1.20 & 1.49 & 1.16 \\ \hline
	I (TL) & 1.26 & 1.17 & 1.29 & \cellcolor{gray!25} --  \\ \hline
	UM & $\mathbf{1.08}$ & $\mathbf{1.00}$ & $\mathbf{1.14}$ & $\mathbf{0.97}$ \\ \hline
\end{tabular}
\caption{\label{table:results}Costs for both experiments, including the standard error (SE), bootstrap (BS), and intrinsic (I) baselines, the same methods with our transfer-learning (TL) calibration, as well as the calibrated constant (C) and \um{}, averaged over the values of $\alpha$ shown in the figures (0.5 to 0.95).}
\end{table}

\section{Ethics Statement}
Training AI models to ``know what they don't know'' is one of the key challenges in AI today~\cite{kindig:2020, Davey:2018, Knight:2018}.
AI models can be notoriously overconfident in scenarios that their training data did not prepare them for~\cite{7298640}, 
eroding trust in AI and society's willingness to accept AI as it continues to replace human decision making in increasingly important roles.
Our work directly addresses this problem, proposing a novel technique for quantifying quantifying model uncertainty that beats all baselines we compare against. 
In addition, we hope that this work will encourage the community to continue to invest and innovate in this important area.
%
%
Due to the nature of this problem we do not foresee any negative consequences stemming from our work.

\bibliography{bibliography}

\clearpage

\section{Appendix to Submission: Learning Prediction Intervals for Model Performance}
\label{sec:supplementary}


\subsection{Performance predictors}
\label{app:method-performance_predictors}

Here we provide further implementation details of the \texttt{meta-model} performance prediction algorithm introduced in Sec.~\ref{sec:method-performance_predictors}, 
as well as the \texttt{crossval} and \texttt{dropout} variants described in Sec.~\ref{sec:methodology-baselines}. 

\paragraph{Meta-model architectures}

The \texttt{meta-model} predictor uses an ensemble of a gbm and a logistic regression model. 
The \texttt{crossval} performance predictor uses an ensemble of ten random forest meta-models, each trained on a separate fold of the test set. 
The \texttt{dropout} predictor uses a single XGBoost~\cite{10.1145/2939672.2939785} regression model (the regression target is the confidence that a base model classification is correct), with the DART~\cite{pmlr-v38-korlakaivinayak15} booster. 
When training the XGBoost model, early stopping is enabled, with the RMSE computed on a holdout set of 20\% of the training data used as the early stopping criteria. 
The dropout regularization was implemented with a dropout rate of 0.25, and zero probability of skipping dropout at each boosting stage. 
The meta-models are all trained with twenty iterations of HPO, using a random search of the parameter spaces described in Table~\ref{table:meta-model-hpo}. 

\begin{table}[!ht]
\centering
  \begin{tabular}{ | l | c | }
    \hline
	Parameter & Range \\ \hline
	\hline
	\multicolumn{2} {| c |} {gbm meta-model} \\ \hline
	$learning\_rate$ & $[0.1, 0.15, 0.2]$ \\ \hline
	$min\_samples\_split$ & $np.linspace(0.005, 0.01, 5)$ \\ \hline
	$min\_samples\_leaf$ &  $np.linspace(0.0005, 0.001, 5)$ \\ \hline
	$max\_leaf\_nodes$ & $[3,5,7,9,11]$ \\ \hline
	$max\_features$ & $[log2, sqrt]$ \\ \hline
	$subsample$ & $[0.3, 0.4, 0.5, 0.6, 0.7, 0.8, 0.9]$ \\ \hline
	$n\_estimators$ & $[100, 150, 200, 250, 300, 350, 400]$ \\ \hline
	\hline
	\multicolumn{2} {| c |} {logistic regression meta-model} \\ \hline
	$C$ & $[0.001, 0.01, 0.1, 1, 10, 100, 1000]$ \\ \hline
	$penalty$ & $[\ell_1, \ell_2]$ \\ \hline
	\hline
	\multicolumn{2} {| c |} {random forest meta-model} \\ \hline
	$n\_estimators$ & $random\_integer \in [5,100]$ \\ \hline
	$max\_depth$ & $[2,3,4,6,None]$ \\ \hline
	$min\_samples\_split$ & $[2,3,4,5,6]$ \\ \hline
	$max\_features$ & $[log2, sqrt]$ \\ \hline 
	\hline
	\multicolumn{2} {| c |} {xgboost meta-model} \\ \hline
	$max\_depth$ & $[2,4,6]$ \\ \hline
	$learning\_rate$ & $[0.05, 0.1, 0.2, 0.3, 0.4, 0.5]$ \\ \hline
	$n\_estimators$ & $[50, 100, 150, 200]$ \\ \hline
	$reg\_lambda$ & $np.logspace(-1, 1, num=20)$ \\ \hline
	$sample\_type$ & $[normal, weighted]$ \\ \hline
	$normalize\_type$ & $[tree, forest]$ \\ \hline
  \end{tabular}
  \caption{\label{table:meta-model-hpo} Hyperparameter ranges included in the random search for HPO performed when fitting the meta-models for performance prediction. }
\end{table}

\paragraph{Features}

The \texttt{meta-model}, \texttt{crossval}, and \texttt{dropout} performance predictors were trained using a set of transformed features designed for the performance prediction problem. 
All three of these predictors used the same features, with the same settings and hyperparameters. 

These transformed features included four derived from the confidence vector from the base model, which forms a probability distribution over the class labels for each data point. 
Three of these features were the highest class confidence, the difference between the highest and second highest class confidence, and the entropy of the confidence vector. 
The last confidence based feature was the proportion of samples in the test set belonging to the class predicted by the base model. 
The intuition behind this feature is that the base model is more likely to make mistakes for classes which were very rare in its training data. 

Three proxy models were used to generate additional features. 
These proxy models (a logistic regression model, a random forest model, and a gbm) were trained on the same train set as the base model, and provide a complementary perspective on the relative classification difficulty of a given data point. 
The random forest and gbm models were trained using fifteen rounds of HPO. 
Once these three proxy models were trained, the top class confidence and 1st - 2nd class confidences from each model were used as features for the meta-model. 

Finally, two additional features were designed to quantify the distance of a given data point from the typical point in the training data. 
The first feature uses the output of the decision function of a one class svm, which classifies samples as either in or out of the training data set. 
To construct the second feature, the "umap kde" transformer first uses the UMAP~\cite{mcinnes2018umap} dimensional reduction algorithm to reduce the dimensionality of the (standard scaled) training dataset down to six dimensions (using parameters $nearest\_neighbors=10$ and $min\_dist=0.1$). 
After dimensional reduction, a separate kernel density estimate (kde) is fit to the data points in each class. 
The kde uses a Gaussian kernel with a bandwidth of 0.2. 
At inference time, each kde provides a probability score for the likelihood that data point was sampled from the distribution described by that kde. 
The highest of these probabilities is returned as the value for the "umap kde" transformed feature.

\paragraph{Calibration}

The meta-models used for performance prediction were all calibrated using a 20\% calibration set held out from their training data (which was the same as the "test set" in the drift scenario). 
Isotonic regression was used to perform the calibration. 
The targets $T_{ir}$ for the isotonic regression calibration were
\begin{equation}
T_{ir} = 
\begin{cases}
\frac{1}{n_- + 2} & \mbox{for $y$ incorrect} \\
\frac{n_+ + 1}{n_+ + 2} & \mbox{for $y$ correct} 
\end{cases}
\, ,
\end{equation}
where $n_+$ and $n_-$ are the number of samples in the calibration set which were correctly and incorrectly predicted by the base model respectively. 
These targets account for finite statistics effects, as they interpolate from equality at $T_{ir} = \frac{1}{2}$ with no calibration samples, to the limit of perfect certainty, where the confidence of a correct classification is zero percent for an incorrectly classified sample, and 100\% for a correctly classified sample, in the limit of infinite calibration samples (assuming a base model accuracy $0 < $ acc $ < 100$).

\paragraph{Performance prediction results}

Table~\ref{table:perfpred-results} presents a comparison of the four performance predictors, in terms of the mean absolute error and the area under the ROC curve for their predictions, for the collection of \ls drift scenarios (left) and \nn drift scenarios (right). 
The inferior performance of the \texttt{crossval} and \texttt{dropout} predictors compared to the \texttt{meta-model} predictor illustrates an additional benefit of the \um{} approach to generating prediction intervals, which is that there is no need to modify the architecture of the meta-model in a potentially detrimental way.

\begin{table}[!ht]
\centering
  \begin{tabular}{ | l | c | c | }
    \hline
	&\multicolumn{2} {| c |} {\ls scenarios}   \\ \hline
	Predictor & MAE & AUC \\ \hline
	\texttt{Confidence} & 4.38 & 0.73   \\ \hline
	\texttt{Crossval} & 4.04 & 0.78  \\ \hline
	\texttt{Dropout} & 4.70 & 0.77  \\ \hline
	\texttt{Meta-model} & $\mathbf{3.84}$ & $\mathbf{0.79}$ \\ \hline
	
    \hline
	& \multicolumn{2} {| c |} {\nn scenarios} \\ \hline
	Predictor & MAE & AUC \\ \hline
	\texttt{Confidence}  & 3.01 & 0.75 \\ \hline
	\texttt{Crossval}  & 2.94 & 0.79 \\ \hline
	\texttt{Dropout} & 3.73 & 0.77 \\ \hline
	\texttt{Meta-model} & $\mathbf{2.78}$ & $\mathbf{0.80}$ \\ \hline
  \end{tabular}

  \caption{\label{table:perfpred-results} Comparison of performance predictor results for \ls drift scenarios (left) and \nn drift scenarios (right). }
\end{table}


\subsection{Uncertainty model}

The \um{} is trained using a quantile loss function, with the target quantile equal to the cost function parameter $\alpha$, and the hyperparameters: $max\_features=log2$, $n\_estimators=200$, and $subsample=0.8$. 
A single hyperparameter configuration was chosen because running HPO to fit the \um{} led to a large increase in variance between random seeds.

\paragraph{Features}

A few additional details are required to fully describe the \um{} features. 
The proxy models for these features were the same proxy models described in App.~\ref{app:method-performance_predictors}. 
The drift models were a set of three random forest models which were trained to distinguish data points in the test set from points in the production set. 
Three of these models were trained which were equivalent except for the set of input features for the model: one drift model was trained on the original feature vectors of the data, one was trained on these features transformed as described in App.~\ref{app:method-performance_predictors}, and the last was trained on the concatenation of the original and transformed feature vectors.

The \texttt{Distance} features all required a histogram, and therefore depend on the choice of histogram bins. 
The histograms for the "top confidence" features from the base model and the three proxy models were binned in units of 0.1 from 0 to 0.9, and in units of 0.01 from 0.9 to 1. 
The top confidences from the drift models tended to exhibit a small set of discrete modes, so a histogram using eleven equally spaced bins, spanning 0 to 1 was used to fully separate these modes. 
The histograms for the "top-2nd confidence" features from the base model and the three proxy models were binned in units of 0.1 from 0 to 1. 
For the "confidence entropy" feature from the base model, the histograms were binned by 0.01 from 0 to 0.1, and by 0.1 from 0.1 to 3. 
The "class frequency" feature from the base model takes categorical values, so it simply has one bin per category. 
Finally, for the "best feature" and "PCA projection" features, the histograms were constructed by choosing ten equally spaced bins spanning the range of the 10-th through 90-th quantiles of the data. 
The outlying data points were not included in the histogram.

The "best feature" dimension was chosen to be the feature dimension in the original data which was assigned the highest feature importance from the random forest proxy model. 
The feature importances from the random forest proxy model were also used to construct the "num important features" feature.

The two "ensemble whitebox" features were designed to capture the level of disagreement between the models in performance predictor's ensemble. The first feature, \texttt{stdev-of-means}, measures how much disagreement there was in each model's average performance.  It is constructed by computing each meta-model's mean performance prediction (averaged over all data points in the production set) and then taking the standard deviation of those means (across meta-models).  The second feature, \texttt{mean-of-stdevs}, captures how much variation there is between models on a per-datapoint level.  It is constructed by computing the standard deviation (across models) of each individual data point's performance predictions, then computing the average (across all data points) of those standard deviations.

The two "calibration whitebox" features are signed and absolute difference in base model accuracy predicted before and after applying the isotonic regression calibration. 

The "gbm whitebox" features summarize information observed within the GBM meta-model.  The first set of features, \texttt{gbm-decision-distance}, were constructed by observing the individual decision trees at prediction time and measuring the distance between each data point's feature value and the decision thresholds it is compared against.  These distances were averaged over all data points in the production set and over all decision trees in the GBM, but were grouped by level in the tree (level 1, level 2, level 3) as well as an aggregate summed across all levels.  

The second set of gbm whitebox features, \texttt{gbm-node-frequencies}, compare how node frequencies of the decision tree traversals change when predicting on test vs production.  It is computed by predicting on the test set and counting the number of times each node in the tree was traversed, then normalizing by the number of data points.  This process is then repeated for the production set to produce two vectors of normalized node frequencies (test and prod).  These vectors are subtracted and the absolute value taken to produce a node-frequency-delta vector.  Features are created by taking the max, average and standard deviation of the node-frequency-delta vector.

   
\section{Experimental methodology}


\subsection{Drift simulation}

\begin{figure}[h]
        \includegraphics[width=0.45\textwidth]{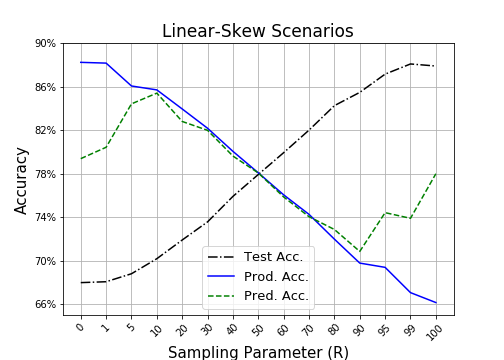}
 \caption{\label{fig:drift-example-a} Example of drift scenarios generated with the \ls methodology, and the predicted accuracy from the \texttt{meta-model} predictor. }
\end{figure}

As an example, Fig.~\ref{fig:drift-example-a} shows the change in base model accuracy, and performance predictions from the \texttt{meta-model} predictor for \ls drift scenarios simulated from the Lending Club dataset. 
Shown are fifteen drift scenarios, derived using the feature $F=$interest rate, using sampling ratios varying from 0 to 100. 
The solid blue (dot-dashed black) curve is the true accuracy of the random forest base model on the production (test) set. 
The gap between the two indicates the change in base model accuracy due to the induced drift. 
The dashed green curve is the \texttt{meta-model} predictor's estimate of the production set accuracy. 
It is easy to see from this plot that the performance prediction works very well until the sampling ratio becomes fairly extreme ($R<5$ or $R>90$). 

\begin{figure}[h]
        \includegraphics[width=0.45\textwidth]{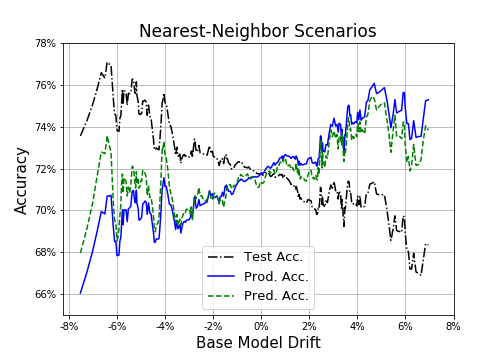}
    \caption{\label{fig:drift-example-b} Example of drift scenarios generated with the \nn methodology, and the predicted accuracy from the \texttt{meta-model} predictor. }
\end{figure}

Fig.~\ref{fig:drift-example-b} shows similar results, for the same dataset and predictor, using drift scenarios generated with the \nn method. 
Shown are the results for 300 scenarios, ordered by amount of change in the base model accuracy (smoothed by a nine-point rolling average).

\subsection{Datasets and settings}

\begin{table*}[t]
\centering
  \begin{tabular}{ | l | c | c | c| r |}
    \hline
    Dataset  & Samples & Features & Classes & Source \\ \hline
    Artificial Character & 10218 & 7 & 10 & {\footnotesize www.openml.org/d/1459} \\ \hline
    Bach Choral & 5,665 & 16 & $7^\dagger$ & {\footnotesize archive.ics.uci.edu/ml/datasets/Bach+Choral+Harmony} \\ \hline
    Bank Marketing  & 38,245 & 16 & 2 & {\footnotesize http://archive.ics.uci.edu/ml/datasets/Bank+Marketing}\\ \hline
    BNG Ionosphere & 100,000* & 34 & 2 & {\footnotesize www.openml.org/d/146} \\ \hline
    BNG Zoo & 100,000* & 16 & 7 & {\footnotesize www.openml.org/d/272} \\ \hline
    Churn Modeling  & 10,000 & 10 & 2& {\footnotesize www.kaggle.com/barelydedicated/bank-customer-churn-modeling}  \\ \hline
    Creditcard Default & 30,000 & $18^\ddagger$ & 2 & {\footnotesize archive.ics.uci.edu/ml/datasets/default+of+credit+card+clients} \\ \hline
    Forest Cover Type & 100,000* & 54 & 7 & {\footnotesize archive.ics.uci.edu/ml/datasets/covertype} \\ \hline
    Higgs Boson & 250,000* & 23 & 2 & {\footnotesize archive.ics.uci.edu/ml/datasets/HIGGS}\\ \hline
    Lending Club Q1 & 78,629 & $19^\ddagger$ & $2^\dagger$ & {\footnotesize www.kaggle.com/wendykan/lending-club-loan-data/data}\\ \hline
    Lending Club Q2 & 118,823 & $19^\ddagger$ & $2^\dagger$ & {\footnotesize www.kaggle.com/wendykan/lending-club-loan-data/data} \\ \hline
    Network Attack & 25,192 & 41 & 2 & {\footnotesize www.kaggle.com/sampadab17/network-intrusion-detection} \\ \hline
    Phishing & 11055 & 30 & 2 & {\footnotesize www.openml.org/d/4534} \\ \hline
    Pulsar & 17.898 & 8 & 2 & {\footnotesize archive.ics.uci.edu/ml/datasets/HTRU2} \\ \hline
    SDSS & 10,000 & $11^\ddagger$ & 3 & {\footnotesize www.kaggle.com/lucidlenn/sloan-digital-sky-survey} \\ \hline
    Waveform & 5000 & 21 & 3 & {\footnotesize archive.ics.uci.edu/ml/datasets/Waveform+Database+Generator} \\ \hline
  \end{tabular}

  \caption{\label{table:datasets} Dataset size information. Some datasets have been modified from their publicly available versions. A * indicates that this dataset was downsampled from its publicly available version. A $\dagger$ indicates that a modified set of class labels was used (as described in the text). }
\end{table*}

Table~\ref{table:datasets} presents more detail about the datasets used in our experimental results.
For computational efficiency, several datasets were downsampled (marked with a *). 
For the Lending Club dataset, only loan records with outcomes of either defaulted or repaid were used, and records which did not yet have a final outcome were excluded. 
Finally the bach choral dataset was preprocessed to combined all classes with the same key (eg G1, G2, . . . G7) into a single label. 
In addition, for several datasets we pruned features as a preprocessing step, in order to improve the data quality ($\ddagger$).



\section{Results}

For completeness, the results from Table~\ref{table:results} are replicated here in Table~\ref{table:results-w-errors}, including the standard errors for all values. 
These standard errors were obtained by bootstrap sampling to construct 95\%-confidence intervals. 
The differences between prediction interval methods within each experiment and performance predictor are all statistically significant, as the confidence intervals do not overlap. 

\begin{table*}[t]
\footnotesize
\centering
\begin{tabular}{| l | c | c | c | c | c | c | c | c |}
	\hline
	&\multicolumn{8} {| c |} {\ls $\rightarrow$ \ls} \\ \hline
	Predictor& SE & BS & I & SE (TL) & BS (TL) & C (TL) & I (TL) & UM \\ \hline
    Confidence & $2.88^{+0.18}_{-0.16}$ & $2.38^{+0.17}_{-0.16}$  & $2.23^{+0.17}_{-0.15}$ & $2.13^{+0.15}_{-0.14}$ & $2.13^{+0.15}_{-0.14}$ & $2.11^{+0.15}_{-0.13}$  & $1.83^{+0.14}_{-0.12}$ & $\mathbf{1.46}^{+0.11}_{-0.09}$ \\ \hline
    Crossval & $2.57^{+0.17}_{-0.15}$ & $2.06^{+0.16}_{-0.14}$ & $1.72^{+0.15}_{-0.13}$ & $1.91^{+0.15}_{-0.13}$ & $1.91^{+0.15}_{-0.13}$ & $1.88^{+0.15}_{-0.13}$ & $1.67^{+0.15}_{-0.12}$ & $\mathbf{1.24}^{+0.11}_{-0.09}$ \\ \hline
    Dropout & $3.09^{+0.18}_{-0.16}$ & $2.50^{+0.18}_{-0.16}$ & $2.20^{+0.17}_{-0.15}$ & $2.13^{+0.16}_{-0.14}$ & $2.13^{+0.16}_{-0.14}$ & $2.10^{+0.15}_{-0.13}$  & $1.86^{+0.15}_{-0.13}$ & $\mathbf{1.38}^{+0.11}_{-0.09}$  \\ \hline
    Meta-model & $2.41^{+0.17}_{-0.15}$ & $1.89^{+0.16}_{-0.14}$ & \cellcolor{gray!25} --  & $1.77^{+0.15}_{-0.13}$ & $1.77^{+0.15}_{-0.13}$ & $1.83^{+0.15}_{-0.13}$  & \cellcolor{gray!25} -- & $\mathbf{1.33}^{+0.12}_{-0.09}$ \\ \hline
	\hline
	&\multicolumn{8} {| c |} {\ls $\rightarrow$ \nn} \\ \hline
	Predictor& SE & BS & I & SE (TL) & BS (TL) & C (TL) & I (TL) & UM \\ \hline
    Confidence & $1.97^{+0.07}_{-0.06}$ & $1.59^{+0.06}_{-0.06}$ & $1.39^{+0.06}_{-0.06}$ & $1.36^{+0.05}_{-0.05}$ & $1.36^{+0.05}_{-0.05}$ & $1.30 ^{+0.05}_{-0.05}$ & $1.26^{+0.05}_{-0.04}$ & $\mathbf{1.08}^{+0.05}_{-0.04}$ \\ \hline
    Crossval & $1.87^{+0.07}_{-0.06}$ & $1.41^{+0.06}_{-0.06}$ & $1.14^{+0.05}_{-0.05}$ & $1.25^{+0.06}_{-0.05}$ & $1.25^{+0.06}_{-0.05}$ & $1.20^{+0.05}_{-0.05}$ & $1.17^{+0.05}_{-0.04}$ & $\mathbf{1.00}^{+0.05}_{-0.04}$ \\ \hline
    Dropout & $2.48^{+0.08}_{-0.08}$ & $2.00^{+0.08}_{-0.07}$  & $1.48^{+0.07}_{-0.06}$ & $1.58^{+0.07}_{-0.06}$ & $1.58^{+0.07}_{-0.06}$ & $1.49^{+0.06}_{-0.06}$  & $1.29^{+0.05}_{-0.05}$ & $\mathbf{1.14}^{+0.05}_{-0.05}$ \\ \hline
    Meta-Model & $1.75^{+0.06}_{-0.06}$ & $1.29^{+0.06}_{-0.06}$ & \cellcolor{gray!25} -- & $1.15^{+0.05}_{-0.05}$ & $1.15^{+0.05}_{-0.05}$ & $1.16^{+0.05}_{-0.05}$ & \cellcolor{gray!25} -- & $\mathbf{0.97}^{+0.04}_{-0.04}$ \\ \hline
\end{tabular}
\caption{\label{table:results-w-errors}Results with standard errors for all experiments, including the standard error (SE), bootstrap (BS), and intrinsic (I) baselines, the same methods with our transfer-learning (TL) calibration, as well as the calibrated constant (C) and \um{} results. These costs are averaged over the ten values of $\alpha$ shown in the Fig.~\ref{fig:biased-target-results} (0.5 to 0.95).}
\end{table*}

\end{document}